\documentclass{article}
\usepackage{ijcai21}

\usepackage{times}
\usepackage{soul}
\usepackage{url}
\usepackage[hidelinks]{hyperref}
\usepackage[utf8]{inputenc}
\usepackage[small]{caption}
\usepackage{graphicx}
\usepackage{amsmath}
\usepackage{amsfonts}
\usepackage{amsthm}
\usepackage{booktabs}
\usepackage{algorithm}
\usepackage{algorithmic}
\usepackage{subcaption}
\DeclareMathOperator*{\argmax}{arg\,max}

\title{Learning Symbolic Expressions via Gumbel-Max Equation Learner Networks}

\author{
    Gang Chen
    \affiliations
    School of Engineering and Computer Science \\ Victoria University of Wellington
    \emails
    aaron.chen@ecs.vuw.ac.nz
}

\begin{document}

\maketitle

\begin{abstract}
Most of the neural networks (NNs) learned via state-of-the-art machine learning techniques are black-box models. For a widespread success of machine learning in science and engineering, it is important to develop new NN architectures to effectively extract high-level mathematical knowledge from complex datasets. Motivated by this understanding, this paper develops a new NN architecture called the Gumbel-Max Equation Learner (GMEQL) network. Different from previously proposed Equation Learner (EQL) networks, GMEQL applies continuous relaxation to the network structure via the Gumbel-Max trick and introduces two types of trainable parameters: structure parameters and regression parameters. This paper also proposes a two-stage training process with new techniques to train structure parameters in both online and offline settings based on an elite repository. On 8 benchmark symbolic regression problems, GMEQL is experimentally shown to outperform several cutting-edge machine learning approaches.
\end{abstract}

\section{Introduction}
\label{sec-int}

Driven by the rising tide of artificial intelligence, machine learning technologies have been increasingly applied to advanced scientific and engineering research \cite{alquraishi2019,goh2017deep}. Although modern machine learning, in particular deep learning, has achieved outstanding success in fulfilling scientists' research demand, most of the neural networks (NNs) learned via these state-of-the-art techniques are \emph{black-box models} \cite{coccia2020}. It is difficult for scientists to directly use such models to drive fruitful scientific explorations. For a widespread success of machine learning in science and engineering, it is important to develop new NN architectures that can effectively extract high-level mathematical knowledge from complex datasets.

Motivated by this research opportunity, innovative machine learning models and methods have been developed recently to solve symbolic regression problems \cite{kim2020,martius2016,sahoo2018,long2018,trask2018}. Symbolic regression aims to induce tractable mathematical expressions, which are expected to reveal valuable scientific insights and are highly interpretable \cite{udrescu2020}. These problems are much more general and difficult than conventional regression problems that often rely on pre-defined and fixed model structures.

Traditionally, symbolic regression is carried out through evolutionary computation methods, including genetic programming (GP) \cite{uy2011,koza1992,back2018}. GP can evolve a tree-based representation of mathematical expressions. It has been successfully utilized to extract fundamental laws of physical systems from experimental data \cite{schmidt2009}. However, as the space of discoverable expressions becomes huge, GP may not scale well and may fail to identify an expression with high precision \cite{kim2020,petersen2019}.

Recently deep learning inspired symbolic regression approaches are gaining attention in the research community \cite{kim2020,sahoo2018,long2018,zheng2018,lu2020,louizos2017}. For example, the \emph{Equation Learner} (EQL) network has been proposed in \cite{martius2016,sahoo2018} to seamlessly integrate symbolic regression and deep learning through a layered network architecture. Each input node in an EQL network represents an individual variable of interest. Other internal nodes represent either a \emph{weighted summation} of all its inputs or an \emph{elementary function}, such as $+$, $-$ and $\times$. All these internal nodes are organized into consecutive layers. Every layer of weighted summation nodes is connected to a subsequent layer of elementary function nodes. Finally, the output of EQL is determined as a weighted summation of all elementary functions in the last hidden layer. To derive a small and easy-to-interpret expression from EQL, weight regularization techniques have been developed in \cite{kim2020,wu2014} to encourage sparse weight distributions across all connections. Based on regularized connections and connection trimming, the final learned expression is formed using all the nodes associated with significant connections.

EQL assumes that every input to an elementary function is a weighted summation. However, for many important mathematical expressions to be learned in practice, such input is determined by either a single variable, constant, or a single elementary function. Moreover, learning regularized connection weights in EQL is highly sensitive to the initial weight settings \cite{kim2020}. Since EQL has many inter-layer connections and hence a large number of connection weights to train, the training process may easily converge to local optima (see empirical evidence in Section \ref{sec-exp}), leading to undesirable regression performance.

To tackle the above issues, we propose a new network architecture called \emph{Gumbel-Max Equation Leaner} (GMEQL) for deep symbolic regression. Different from EQL, all internal nodes in GMEQL represent elementary functions and are organized into multiple layers. Each input of an \emph{elementary function node} is connected to the output of another node in the previous layer. This connection is \emph{stochastic}. We define a \emph{connection instance} as a link from the output of a specific node in one layer to the specific input of another node in the next layer. Every connection is hence associated with multiple connection instances and each connection instance can be sampled with a certain probability. After sampling one connection instance for every connection in GMEQL, a mathematical expression can be obtained through either depth-first or breadth-first traversal of GMEQL, starting from the output nodes.

We can identify two types of trainable parameters in GMEQL: \emph{structure parameters} that govern the probabilities of sampling any connection instances; and \emph{regression parameters} that control the weights of all connections. Note that all connection instances for the same connection share the same weight. Hence the total number of regression parameters is much smaller in GMEQL, compared to EQL. Typically, optimizing structure parameters requires us to sample many connection instances for each connection and evaluate the benefit of using every sampled connection instance \cite{schulman2015,williams1992}. Since this sampling process is performed in a discrete manner, it is difficult to train structure parameters directly through backpropagation.

For effective training of structure parameters, we propose to apply \emph{continuous relaxation} to all connection instances based on the \emph{Gumbel-Max} reparameterization trick \cite{maddison2016,kingma2013,jimenez2014}. Consequently, the \emph{degree of involvement} of each connection instance in GMEQL can be measured continuously. This makes it straightforward to train structure parameters through backpropagation. We notice that structure parameters and regression parameters depend heavily on each other. Improper handling of this interdependence can lead to poor training performance. To address this issue, we develop a \emph{two-stage training process} with \emph{stage 1} focusing on training structure parameters alone. Meanwhile, to increase the chance of discovering desirable network structures, we propose to use an \emph{elite repository} to guide the sampling of connection instances in GMEQL and to facilitate structure parameter training in both online and offline settings.

On 8 benchmark symbolic regression problems, we experimentally compare GMEQL, EQL and another deep reinforcement learning powered algorithm called \emph{deep symbolic regression} {DSR} \cite{petersen2019}. GMEQL is also compared to the standard GP method. Our experiments clearly show that GMEQL can achieve the highest precision on most of the evaluated symbolic regression tasks.

\section{Related Work}
\label{sec-rw}

Significant efforts have been made in recent years to design NN architectures that can extract interpretable knowledge required for scientific explorations. For example, EQL is developed in \cite{martius2016,sahoo2018} to induce mathematical expressions from datasets. While being successful on a range of benchmark problems, the performance of EQL can be significantly enhanced by addressing several major issues discussed in Section \ref{sec-int}. Another deep learning architecture called the PDE-Net uses constrained convolution operators to identify differential operators that can accurately predict the dynamics of spatiotemporal systems \cite{long2018}. It is possible to incorporate such differential operators as activation functions in EQL and GMEQL. However, following many existing works \cite{kim2020,petersen2019,uy2011}, this paper focuses on learning conventional expressions.

Symbolic regression can also be approached through building a generative system \cite{petersen2019,kusner2017}. For example, GrammarVAE is developed in \cite{kusner2017} to generate discrete structures. However this model does not guarantee to generate syntactically valid expressions. Recurrent neural network (RNN) stands for another powerful generative model \cite{lipton2015}. For instance, DSR proposed in \cite{petersen2019} can train a RNN policy via reinforcement learning to generate descriptive expressions. DSR has successfully solved many benchmark symbolic regression problems. An interactive visualization platform has also been developed to facilitate its practical use \cite{kim20201}. Rather than searching for mathematical expressions indirectly through RNNs and reinforcement learning, it may be more straightforward and efficient to directly explore and optimize expressions embedded within an expressive NN architecture such as GMEQL. In Section \ref{sec-exp}, the effectiveness of GMEQL and DSR will be studied experimentally.

Our proposed use of the Gumbel-Max trick in GMEQL is closely related to several existing methods for neural architecture search (NAS) \cite{zoph2016,liu2018,xie2018}. Particularly, the stochastic NAS (SNAS) algorithm exploits this trick to optimize all connections within every cell of a large NN through gradient-descent search \cite{xie2018}. Different from SNAS, we extend EQL with the Gumbel-Max trick to facilitate gradient-based learning of mathematical expressions. Furthermore, to improve training performance and expedite the search for accurate expressions, new techniques will be developed in Section \ref{sec-alg} to train structure parameters.

\section{Gumbel-Max EQL Network}
\label{sec-arc}

\begin{figure}[!ht]
\includegraphics[width=0.9\columnwidth]{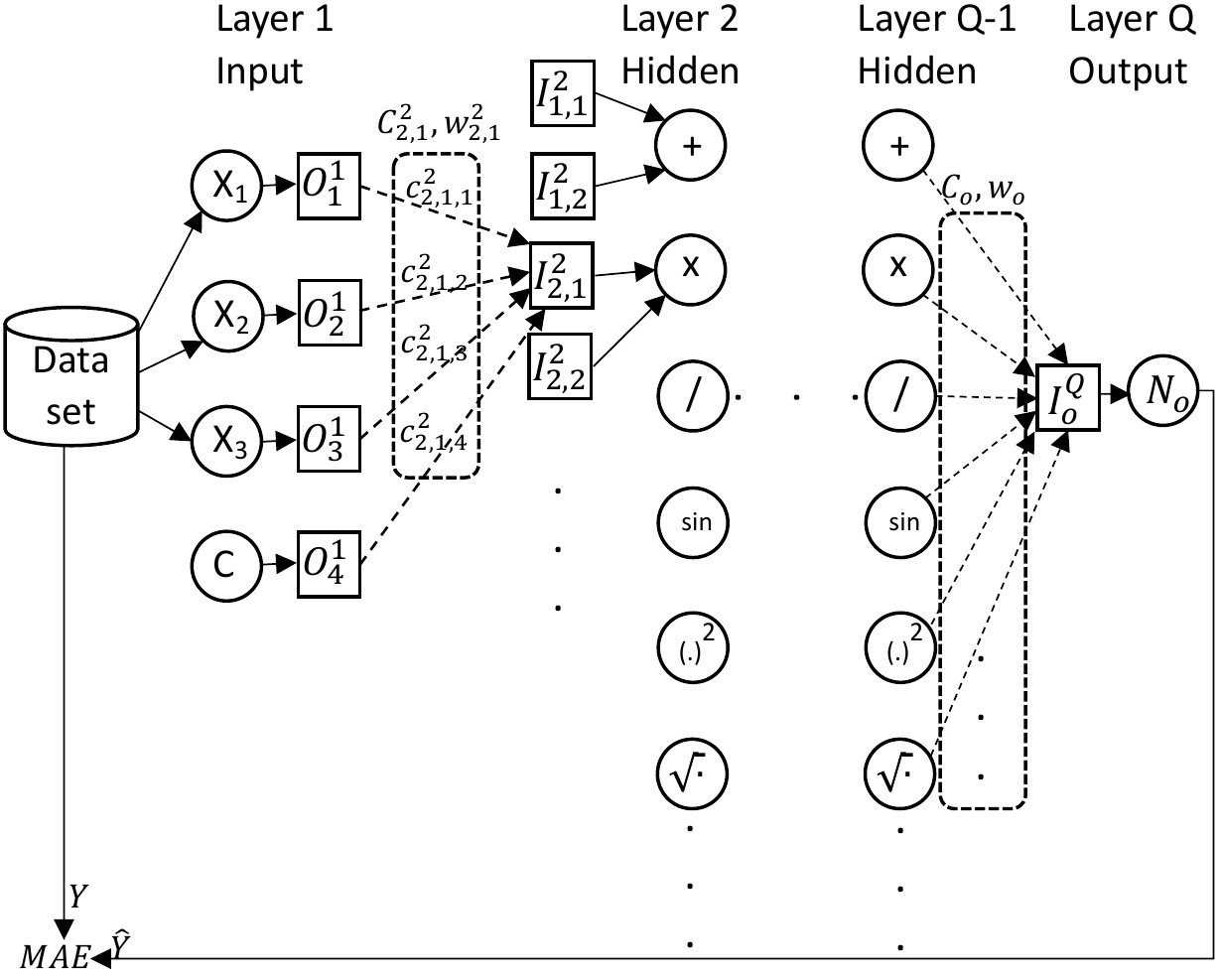}
\caption{A GMEQL network example for symbolic regression.}
\label{fig-gmeql}
\end{figure}

The architecture of the newly proposed GMEQL network is depicted in Figure \ref{fig-gmeql}. The network comprises of multiple layers, including one input layer, several hidden layers and one output layer. Each node in the input layer represents either a variable of interest in the regression dataset or a constant (with default value of 1). Each node in the output layer denotes a target output included in the same dataset. Since many regression problems have a single target output \cite{uy2011,udrescu2020}, only one output node is illustrated in Figure \ref{fig-gmeql}.

The hidden layers occupy the main body of GMEQL. Each hidden layer contains multiple hidden nodes with each node representing an elementary function. Choosing suitable elementary functions for every hidden node is a domain-dependent task. In addition to $+$, $\times$ and $/$, other functions can also be employed for building a GMEQL network. In our experiments, we manually select a group of elementary functions commonly used for symbolic regression, including $+$, $-$, $\times$, $/$, $sin(\cdot)$, $cos(\cdot)$, $\sqrt{\cdot}$, $(\cdot)^{2\sim 6}$, and $\sum$. Future research will explore the possibility of automatically choosing/designing elementary functions for GMEQL.

Each hidden \emph{function node} in Figure \ref{fig-gmeql} can have a varied number of inputs. Denote the $j$-th input of node $N_i^k$ (i.e., the $i$-th node in the $k$-th layer) as $I_{i,j}^k$. The corresponding connection $C_{i,j}^k$ of input $I_{i,j}^k$ is defined as a finite set of connection instances:
$$
C_{i,j}^k = \{ c_{i,j,l}^k \}_{l=1}^{M_{k-1}},
$$
where $c_{i,j,l}^k$ refers to the $l$-th connection instance. It links the output of node $N_l^{k-1}$ in the $(k-1)$-th layer to the input $I_{i,j}^k$ of node $N_i^k$ in the $k$-th layer. Hence the total number of connection instances for any connection $C_{i,j}^k$ in the $k$-th layer equals to the size of the $(k-1)$-th layer, i.e., $M_{k-1}$. Upon extracting a mathematical expression from GMEQL, we must sample exactly one connection instance for each connection. Any sampling decision for connection $C_{i,j}^k$ can be represented as an \emph{one-hot vector} $S_{i,j}^k=[1,0,\ldots]^T$. Each dimension of $S_{i,j}^k$ corresponds to a separate connection instance of $C_{i,j}^k$. Accordingly, the $j$-th input of node $N_i^k$ can be determined as:
$$
I_{i,j}^k = w_{i,j}^k \left( [O_1^{k-1},\ldots,O_{M_{k-1}}^{k-1}] \cdot S_{i,j}^k \right),
$$
where $O_l^{k-1}$ gives the output of node $N_l^{k-1}$ in the $(k-1)$-th layer. All connection instances of connection $C_{i,j}^k$ share the same scalar weight factor $w_{i,j}^k$ in the above equation. Based on all its input, the output of node $N_i^k$ can be computed as:
$$
O_i^k = N_i^k(I_{i,1}^k,I_{i,2}^k,\ldots),
$$
with $N_i^k$ operating as a pre-defined elementary function. Different from hidden nodes, an output node has a single input and its output is set identical to its input.

Considering a GMEQL network with a single output node $N_o$ (see Figure \ref{fig-gmeql}), based on any variable inputs $X$ to GMEQL obtained from a dataset, the network output $\hat{Y}$ depends on the one-hot vector $S_{i,j}^k$ for each connection $C_{i,j}^k$ as well as all the connection weights $\{w_{i,j}^k\}$. This input-output relationship is captured by function $\hat{Y}(X, \{S_{i,j}^k\}, \{w_{i,j}^k\})$. For any dataset with $R$ data instances $\{X_i, Y_i\}_{i=1}^R$, the \emph{mean absolute error} (MAE) of GMEQL on this dataset is defined as:
\begin{equation}
MAE = \frac{1}{R} \sum_{i=1}^R \left| Y_i - \hat{Y}(X_i, \{S_{i,j}^k\}, \{w_{i,j}^k\})  \right|
\label{equ-mae}
\end{equation}
The \emph{goal} of symbolic regression is to minimize MAE by identifying optimal $\{S_{i,j}^k\}$ and $\{w_{i,j}^k\}$.

Optimizing $\{w_{i,j}^k\}$ can be approached directly through backpropagation training guided by MAE. Optimizing $\{S_{i,j}^k\}$ is conducted in the discrete space of all one-hot vectors and is more challenging. To effectively optimize $\{S_{i,j}^k\}$, we adopt the Gumbel-Max trick by introducing the \emph{structure parameters} $Z_{i,j}^k=\{z_{i,j,l}^k\}_{l=1}^{M_{k-1}}$ with respect to each connection $C_{i,j}^k$. The connection weight $w_{i,j}^k$ for the same connection will be termed the \emph{regression parameters}. Using $Z_{i,j}^k$, continuous relaxation can be applied to $S_{i,j}^k$. Specifically $V_{i,j}^k$ in (\ref{equ-vij}) measures the \emph{degree of involvement} of each connection instance. It replaces $S_{i,j}^k$ in $\hat{Y}(X, \{S_{i,j}^k\}, \{w_{i,j}^k\})$ while training GMEQL.
\begin{equation}
V_{i,j}^k = \sigma\left( \left[\frac{z_{i,j,1}^k + G_{i,j,1}^k}{\lambda},\ldots,\frac{z_{i,j,M_{k-1}}^k + G_{i,j,M_{k-1}}^k}{\lambda} \right]^T \right).
\label{equ-vij}
\end{equation}
Here $\sigma(\cdot)$ is the standard \emph{softmax} function. $G_{i,j,l}^k$, $1\leq l\leq M_{k-1}$, are independently sampled Gumbel random variables and $\lambda>0$ is the temperature parameter. When $\lambda$ is small, randomly generated vector $V_{i,j}^k$ in (\ref{equ-vij}) is expected to be very close to an one-hot vector. Using \cite{maddison2016} (see Appendix A), the probability for the $l$-th dimension of $V_{i,j}^k$, i.e. $V_{i,j}^k(l)$, to be 1 (or close to 1) is
$$
\mathbb{P}(V_{i,j}^k(l)\approx 1)=\frac{\exp(z_{i,j,l}^k)}{\sum_{l=1}^{M_{k-1}} \exp(z_{i,j,l}^k) }.
$$
It is immediate to see that structure parameters $\{Z_{i,j}^k\}$ in GMEQL control the sampling of one-hot like vectors for all connections. Because MAE in (\ref{equ-mae}) is differentiable with respect to $\{Z_{i,j}^k\}$, structure parameters can be trained using gradient-descent methods, the same as regression parameters.

\section{Network Training Algorithm}
\label{sec-alg}

Although a GMEQL network can be trained directly through backpropagation, we must address three key issues in order to achieve high training performance. The {\bf first issue} is caused by the strong interdependence between structure and regression parameters. We found through experiments that jointly training both the structure and regression parameters immediately after network initialization can lead to inferior performance. In fact, it takes time for GMEQL to discover accurate network structures through repeated sampling of connection instances for all connections. This sampling process is strongly influenced by the regression parameters. If the regression parameters are heavily trained in the initial training stage, future sampling of connection instances can be seriously biased.

The {\bf second issue} concerns the accuracy of the sampled connection instances. In order for GMEQL to learn complex expressions accurately, we must correctly sample connection instances for many connections. The joint probability of doing so can be very low, even after training GMEQL for many iterations.

The {\bf third issue} considers the question of how to re-use previously sampled networks to expedite the training process. Existing training techniques are developed mainly for the online setting \cite{kim2020,petersen2019,sahoo2018}. During each training iteration, numerous new samples of connection instances must be generated to train structure parameters (more discussion in Subsection \ref{sub-sec-er}). To improve sample efficiency, it is desirable to train structure parameters in the offline setting based on accurate network samples obtained previously.

\subsection{Two-Stage Training Process}
\label{sub-sec-tst}

To address the first issue, we develop a two-stage training process as summarized in Algorithm \ref{alg-train}. Stage 1 focuses solely on training structure parameters in the online setting. By doing so, we can mitigate the bias caused by training regression parameters concurrently. To ease discussion, we define the set of randomly generated vectors $\{V_{i,j}^k\}$ in (\ref{equ-vij}) with respect to all connections of GMEQL as a \emph{sampled network instance}, indicated as $T$. The network instance with the highest accuracy ever sampled is indicated as $T^*$. 

In order to discover as many accurate network instances as possible, we must properly initialize all structure parameters $\{Z_{i,j}^k\}$ and regression parameters $\{w_{i,j}^k\}$\footnote{In our experiments, these parameters are initialized independently based on the standard normal distribution}. To reduce the bias imposed by randomly initialized parameters, multiple rounds of independent training of structure parameters will be carried out in stage 1. The most accurate network instances identified across all these rounds will be utilized to train both structure and regression parameters in stage 2.

\begin{algorithm}[!ht]
\begin{algorithmic}
\STATE {\bf Input}: GMEQL network, elite repository $\Delta$, learning rate
\STATE {\bf Output}: $T^*$
\STATE {\bf Stage 1}
\STATE \ \ Repeat for $n$ rounds:
\STATE \ \ \ \ Initialize $\{Z_{i,j}^k\}$ and $\{w_{i,j}^k\}$
\STATE \ \ \ \ Repeat for $m$ rounds:
\STATE \ \ \ \ \ \ Sample $p$ network instances: $T_1,\ldots,T_p$
\STATE \ \ \ \ \ \ Evaluate MAE of $T_1,\ldots,T_p$
\STATE \ \ \ \ \ \ Store $T_1,\ldots,T_p$ in $\Delta$ according to MAE
\STATE \ \ \ \ \ \ Train $\{Z_{i,j}^k\}$ to minimize the average MAE
\STATE \ \ \ \ \ \ \ \ across $T_1,\ldots,T_p$ in the online setting
\STATE {\bf Stage 2}
\STATE \ \ Initialize $\{Z_{i,j}^k\}$ and $\{w_{i,j}^k\}$
\STATE \ \ Repeat for $q$ rounds:
\STATE \ \ \ \ Sample $p$ network instances: $T_1,\ldots,T_p$
\STATE \ \ \ \ Evaluate MAE of $T_1,\ldots,T_p$
\STATE \ \ \ \ Store $T_1,\ldots,T_p$ in $\Delta$ according to MAE
\STATE \ \ \ \ Train $\{Z_{i,j}^k\}$ and $\{w_{i,j}^k\}$ to minimize the average MAE
\STATE \ \ \ \ \ \ across $T_1,\ldots,T_p$ in the online setting
\STATE \ \ \ \ Random sample $r$ network instances from $\Delta$
\STATE \ \ \ \ Train $\{Z_{i,j}^k\}$ in the offline setting using the r instances
\end{algorithmic}
\caption{The algorithm for training GMEQL.}
\label{alg-train}
\end{algorithm}
All trainable parameters will be re-initialized at the beginning of stage 2. This is experimentally shown to improve training stability slightly. Hence, only accurate network instances discovered during stage 1 will be passed to stage 2 through the elite repository $\Delta$. Different from stage 1, both $\{Z_{i,j}^k\}$ and $\{w_{i,j}^k\}$ are trained in the online setting in stage 2. $\{Z_{i,j}^k\}$ will be trained in the offline setting too.

\subsection{Elite Repository}
\label{sub-sec-er}

To address the second issue, we build an elite repository $\Delta$ in Algorithm \ref{alg-train} to guide the sampling of network instances. Each network instance stored in $\Delta$ is associated with its own MAE evaluated through (\ref{equ-mae}). $\Delta$ has a fixed size and only keeps those sampled instances with the lowest MAE. To promote diversity, every instance is converted to an expression through breadth-first graph traversal from the output node, ignoring all connection weights. We require any two network instances in $\Delta$ to produce different expressions.

Before sampling a new network instance $T$, an existing instance from $\Delta$ will be randomly selected. All instances in $\Delta$ are sorted incrementally based on MAE. One instance $\tilde{T}$ in the sorted list is chosen according to the \emph{power-law distribution} with exponent 1.5 \cite{clauset2009}. $\tilde{T}$ is copied to produce $T$. Afterwards the vectors $V_{i,j}^k$ for 20\%\footnote{We experimentally evaluated other settings such as 10\% and 30\% and found that 20\% can produce slightly better results.} of the connections in $T$ are re-sampled using (\ref{equ-vij}). By only sampling a small portion of $V_{i,j}^k$ using structure parameters and inheriting the rest from an accurate network instance $\tilde{T}\in\Delta$, we can significantly increase the chance for Algorithm \ref{alg-train} to discover precise network instances (see experiment results in Section \ref{sec-exp}).

We label all re-sampled connections in $T$. During online training of $\{Z_{i,j}^k\}$, only those structure parameters associated with labelled connections will be updated to minimize MAE. This is important to mitigate the bias introduced by $\tilde{T}$. During each training iteration, Algorithm \ref{alg-train} samples $p$ new network instances. Driven by the average MAE across all the $p$ instances, $\{Z_{i,j}^k\}$ and $\{w_{i,j}^k\}$ can be trained to reduce MAE through backpropagation. Because training is performed on network instances sampled according to $\{Z_{i,j}^k\}$, it is termed \emph{online training} in this paper. $\{Z_{i,j}^k\}$ can also be trained using previously sampled network instances (these instances were not sampled according to the current $\{Z_{i,j}^k\}$). The corresponding \emph{offline training} technique will be developed in the next subsection.

\subsection{Offline Training of Structure Parameters}
\label{sub-sec-offt}

To tackle the third issue, a new technique to train $\{Z_{i,j}^k\}$ in the offline setting is developed in this subsection. Define
\begin{equation}
J=\sum_{T} \mathbb{P}_{\{Z_{i,j}^k\}}(T) \frac{1}{MAE(T)}
\label{equ-j}
\end{equation}
as the \emph{expected sampling performance} of structure parameters $\{Z_{i,j}^k\}$. $\mathbb{P}$ indicates the probability of sampling any network instance $T$ according to $\{Z_{i,j}^k\}$. We use the reciprocal of MAE achievable by $T$ in (\ref{equ-j}) to set the target of maximizing $J$. Considering a different distribution $\mathbb{Q}$ for sampling $T$, it is easy to see that
\begin{equation}
\begin{split}
\log J & = \log \sum_{T} \frac{\mathbb{Q}(T)}{\mathbb{Q}(T)} \mathbb{P}_{\{Z_{i,j}^k\}}(T) \frac{1}{MAE(T)} \\
& \geq \sum_{T} \mathbb{Q}(T) \log \frac{\mathbb{P}_{\{Z_{i,j}^k\}}(T)}{\mathbb{Q}(T)}\frac{1}{MAE(T)}=L.
\end{split}
\label{equ-j-bound}
\end{equation}
Rather than maximizing $J$ directly, its lower bound $L$ in (\ref{equ-j-bound}) can be maximized. In fact,
\begin{equation}
\argmax_{\{Z_{i,j}^k\}} L = \argmax_{\{Z_{i,j}^k\}} \sum_{T} \mathbb{Q}(T) \log \mathbb{P}_{\{Z_{i,j}^k\}}(T).
\label{equ-max-l}
\end{equation}
Assume $\mathbb{Q}$ is realized by sampling network instances from $\Delta$ according to a power-law distribution. Given a batch $\mathcal{B}$ of network instances retrieved from $\Delta$, $\{Z_{i,j}^k\}$ can be trained in the offline setting with the gradient estimated below:
\begin{equation}
\frac{1}{\|\mathcal{B}\|}\sum_{T\in\mathcal{B}} \nabla_{\{Z_{i,j}^k\}} \log \mathbb{P}_{\{Z_{i,j}^k\}}(T).
\label{equ-g}
\end{equation}
Consider any network instance $T$ in (\ref{equ-g}). Let $V_{i,j}^k$ be the one-hot like vector with respect to connection $C_{i,j}^k$ of $T$, which can be obtained from (\ref{equ-vij}) with a certain probability. The equation below can be derived using \cite{maddison2016} (see Appendix B):
\begin{equation}
\begin{split}
& \nabla_{z_{i,j,l}^k} \log \mathbb{P}_{\{Z_{i,j}^k\}}(T)= \\
& \sum_{a=1}^{M_{k-1}} \nabla_{z_{i,j,l}^k} \log\left( \frac{\exp(z_{i,j,a}^k) (V_{i,j}^k(a))^{-\lambda-1}}{\sum_{b=1}^{M_{k-1}} \exp(z_{i,j,b}^k) (V_{i,j}^k(b))^{-\lambda}} \right) \\
& = \frac{\sum_{a=1,a\neq l}^{M_{k-1}} \exp(z_{i,j,a}^k) (V_{i,j}^k(a))^{-\lambda}}{\sum_{b=1}^{M_{k-1}} \exp(z_{i,j,b}^k) (V_{i,j}^k(b))^{-\lambda}} - \\
& \ \ \ \ \ \frac{(M_{k-1}-1) \exp(z_{i,j,l}^k) (V_{i,j}^k(l))^{-\lambda}}{\sum_{b=1}^{M_{k-1}} \exp(z_{i,j,b}^k) (V_{i,j}^k(b))^{-\lambda}}
\end{split}
\label{equ-grad-form}
\end{equation}
Using the gradients calculated above, structure parameters can be trained offline based on accurate network instances sampled previously, without incurring extra sample cost.

\section{Experiment}
\label{sec-exp}

GMEQL, EQL, DSR and a standard GP method are experimentally evaluated in this section on 8 benchmark problems, which have been summarized in Table \ref{tab-ben}. In this table, problems $b1$ to $b6$ are derived from similar problems in many existing works \cite{petersen2019,kim2020,uy2011}. Different from the original problem versions, we increase the number of variables and introduce multiple real-valued coefficients to make these problems more challenging. Problems $b7$ and $b8$ are derived from important physics formula studied in \cite{udrescu2020}. Problem $b8$ contains both $/$ and $\sqrt{\cdot}$ as two component functions. We study this problem because it was shown previously that $/$ and $\sqrt{\cdot}$ cannot be handled easily by a NN \cite{sahoo2018}.
\begin{table}[htb!]
\caption{Benchmark symbolic regression functions.}
\resizebox{\columnwidth}{!}{
\centering
\begin{tabular}{||l|c|c|c||}
\hline
ID & Expression & No. of & Variable \\
   &            & Examples & Value Range \\
\hline
$b1$ & $0.8 x_1^3 + 0.9 x_2^2 + 1.2 x_3$ & 300  & $(0,2)^3$ \\
$b2$ & $0.8 x_3^4 + 0.8 x_1^3 + 1.2 x_2^2 + 1.4 x_3 $ & 300 & $(0,2)^3$ \\
$b3$ & $0.8 x_3^5+ 1.2 x_2^4 + 1.2 x_1^3 + 0.9 x_2^2 + 1.1 x_3$ & 300 & $(0,2)^3$ \\
$b4$ & $1.1 x_1^6 + 0.8 x_2^5 + 0.9 x_3^4 + 1.3 x_1^3 + 1.2 x_2^2$ & 300 & $(0,2)^3$ \\
     & $+ 0.9 x_3$ &   & \\
$b5$ & $1.5 sin(x_1) + 1.3 sin(x_2^2)$ & 300 & $(0,10)^2$ \\
$b6$ & $1.2 sin(1.1 x_1) cos(0.9 x_2)$ & 300 & $(0,10)^2$ \\
$b7$ & $1.1 x_1 + 0.9 x_2 + 2.1 x_1 x_2 cos(1.2 x_3)$ & 300 & $(0,10)^3$ \\
$b8$ & $\sqrt{1.3 + 1.2 \frac{x_2}{x_1}}$ & 300 & $(1,20)^3$ \\
\hline
\end{tabular}
}
\label{tab-ben}
\end{table}
\begin{figure*}[!ht]
\begin{minipage}[t]{0.24\textwidth}
\includegraphics[width=\textwidth]{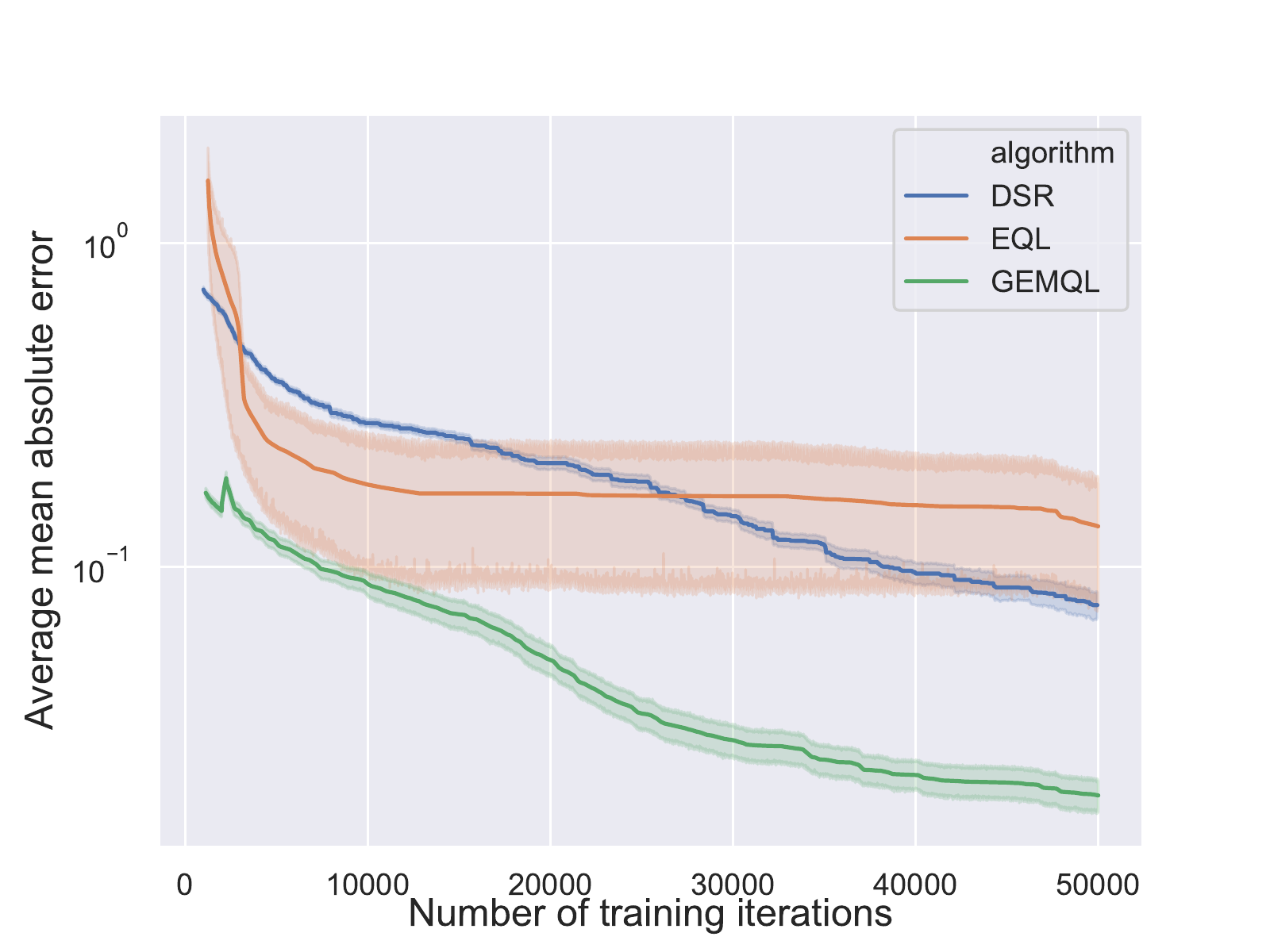}
\subcaption{$b1$}
\end{minipage}
\begin{minipage}[t]{0.24\textwidth}
\includegraphics[width=\textwidth]{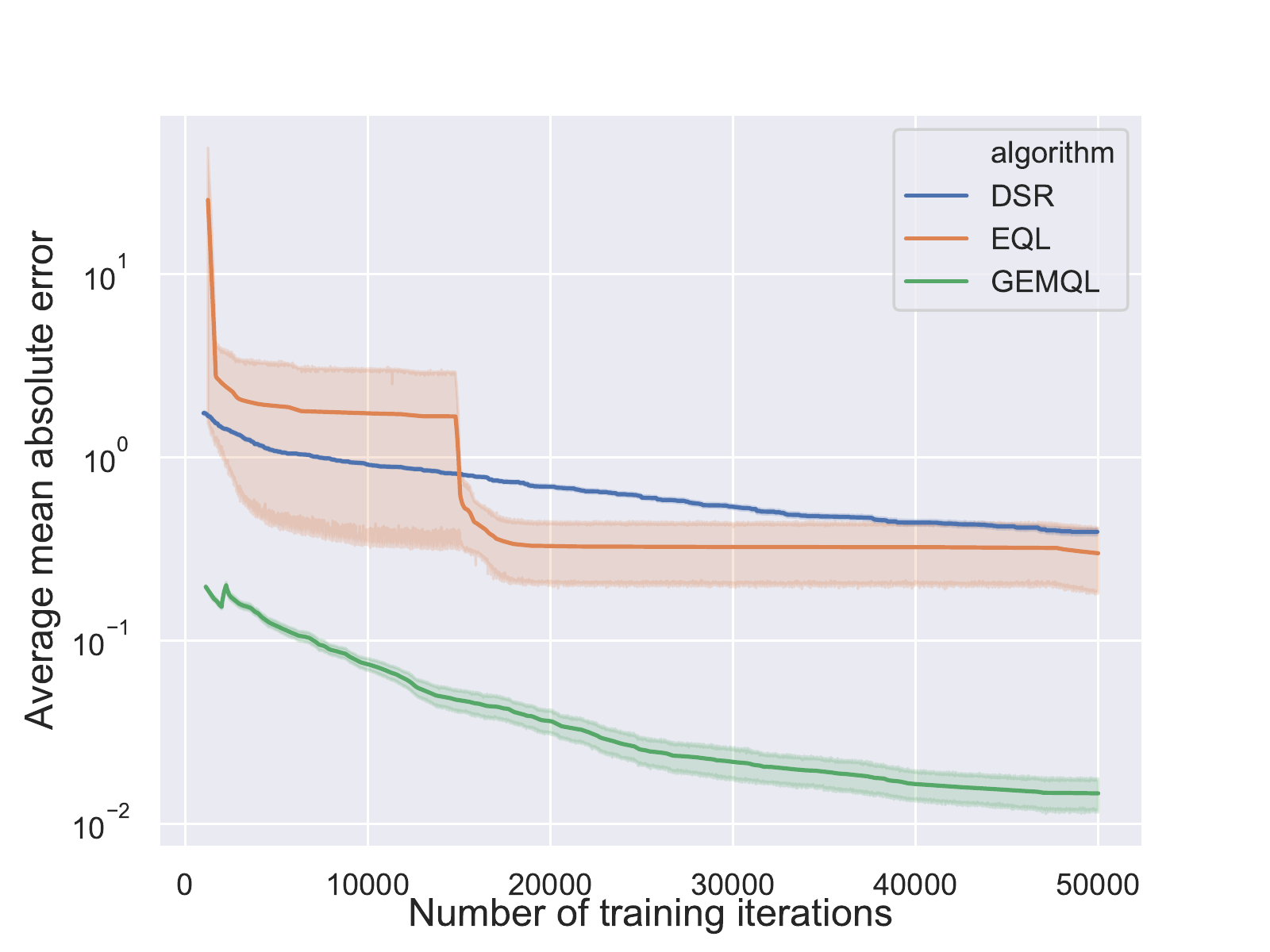}
\subcaption{$b2$}
\end{minipage}
\begin{minipage}[t]{0.24\textwidth}
\includegraphics[width=\textwidth]{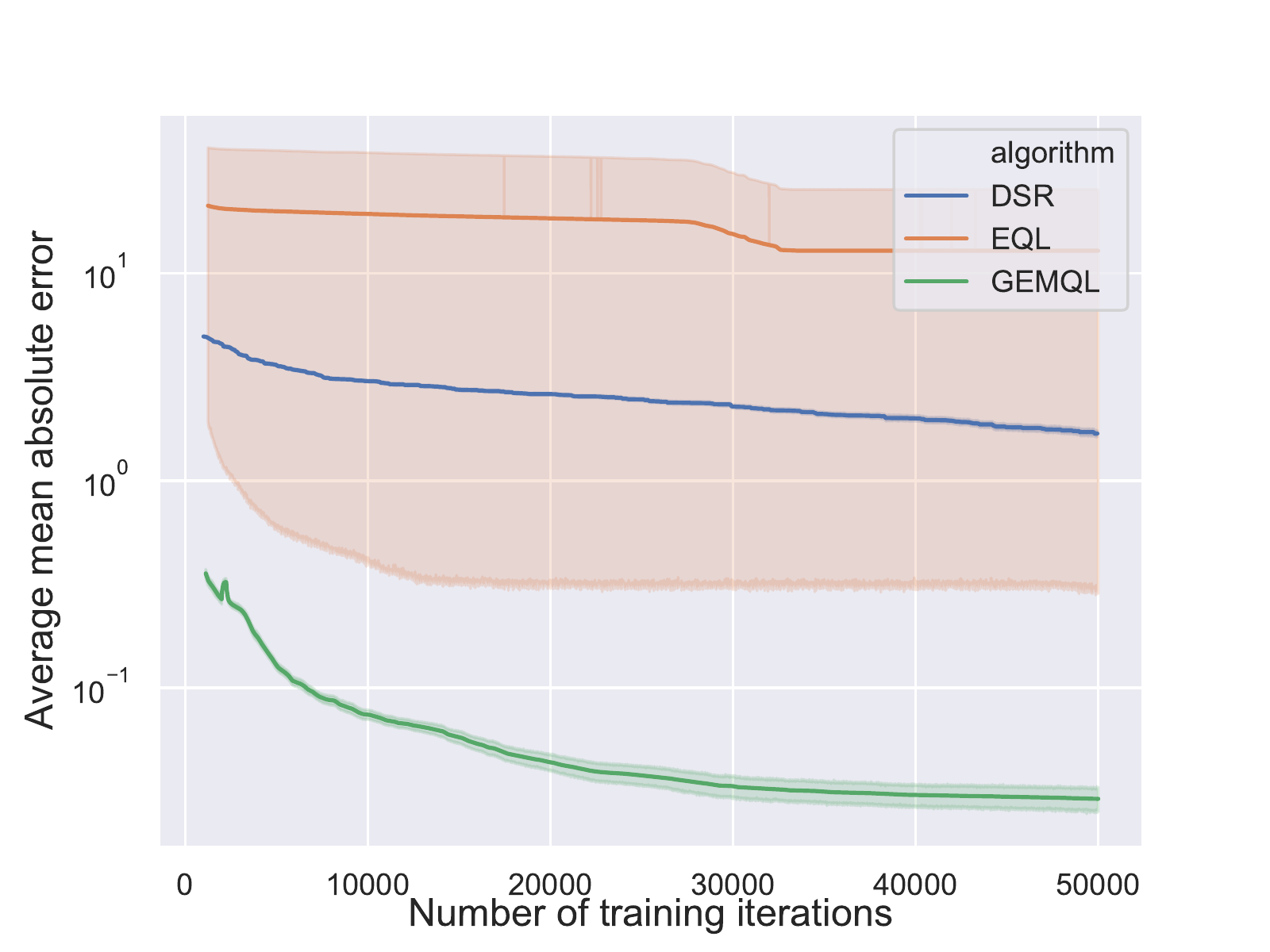}
\subcaption{$b3$}
\end{minipage}
\begin{minipage}[t]{0.24\textwidth}
\includegraphics[width=\textwidth]{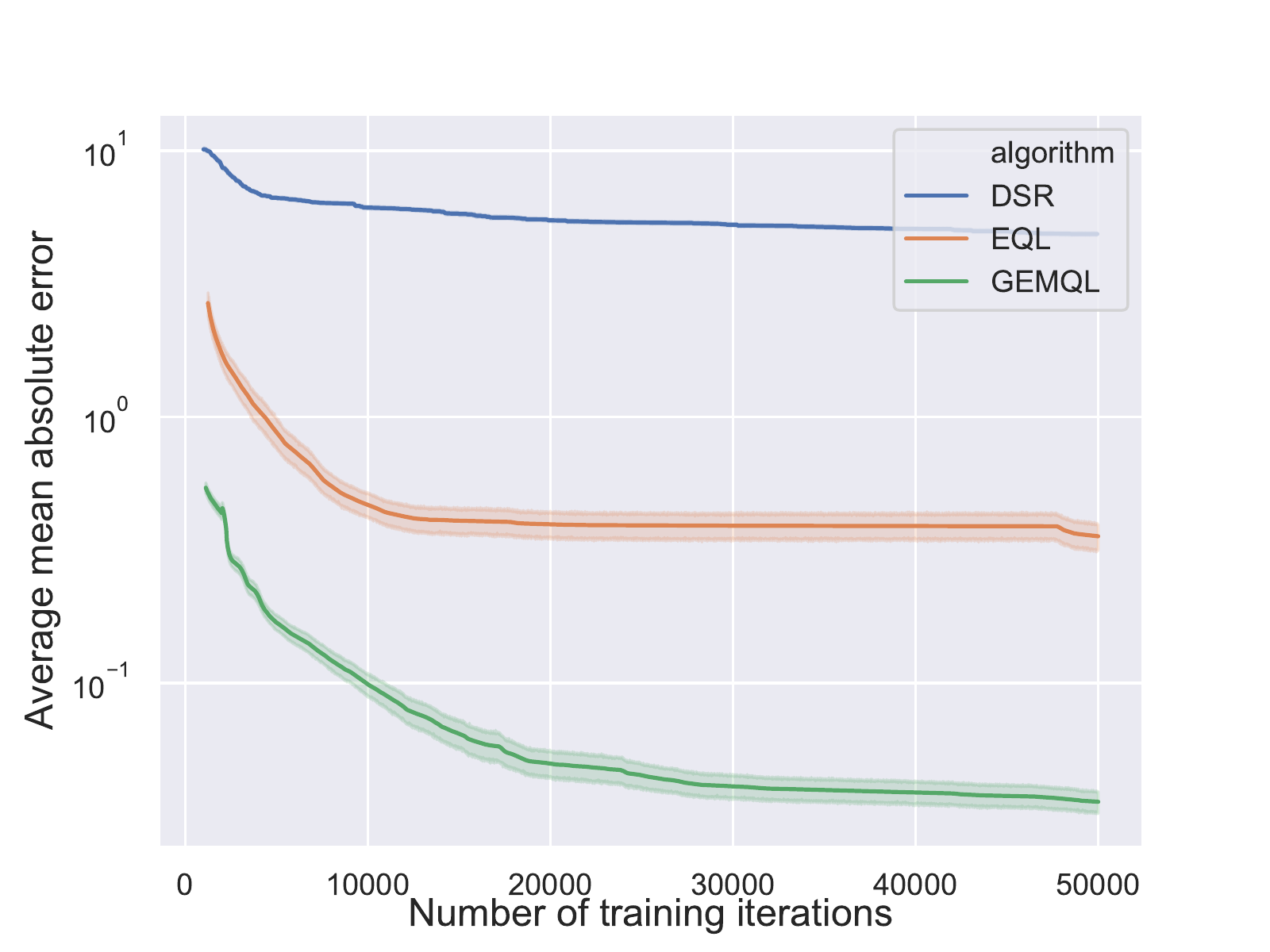}
\subcaption{$b4$}
\end{minipage}
\\
\begin{minipage}[t]{0.24\textwidth}
\includegraphics[width=\textwidth]{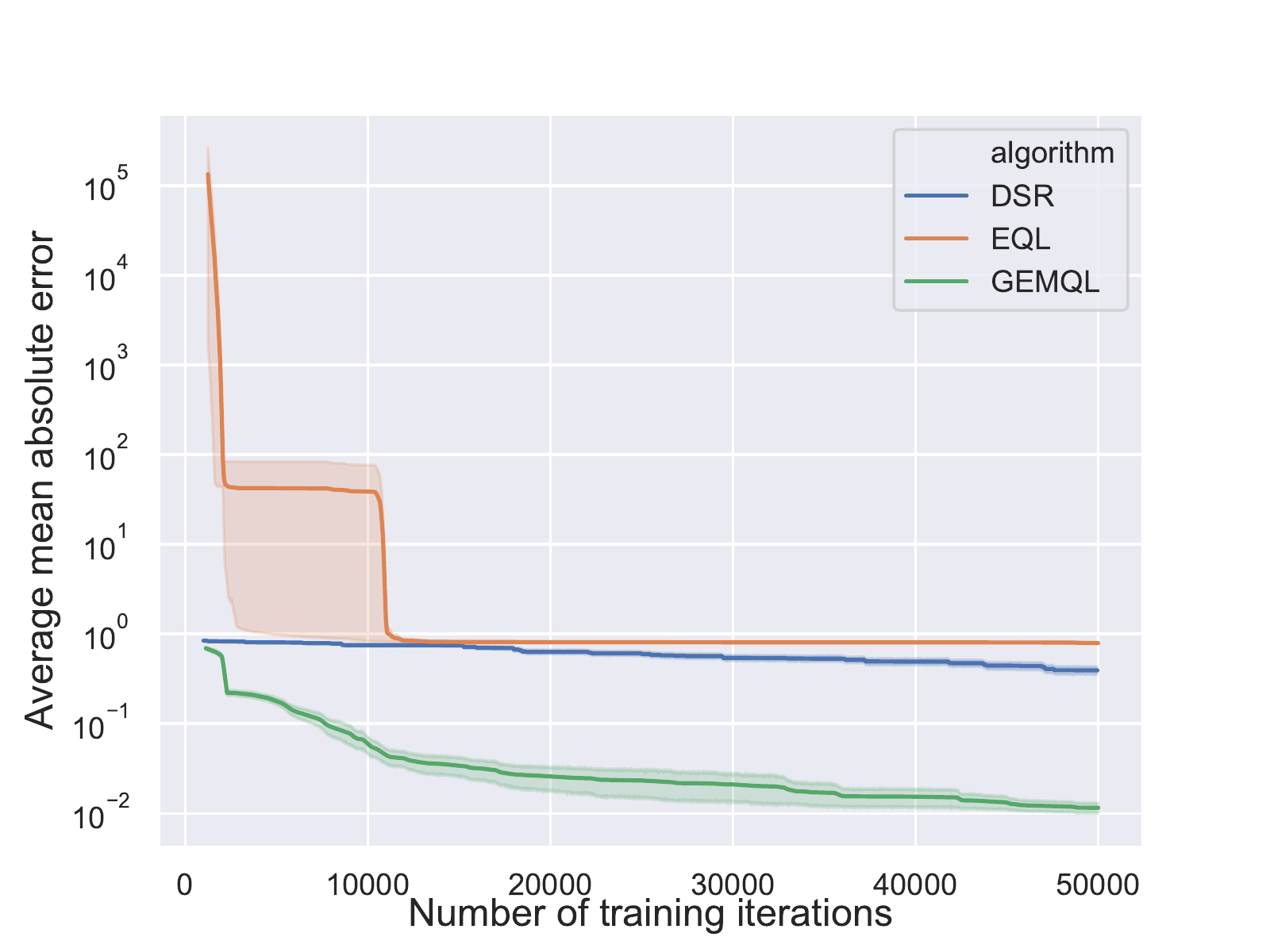}
\subcaption{$b5$}
\end{minipage}
\begin{minipage}[t]{0.24\textwidth}
\includegraphics[width=\textwidth]{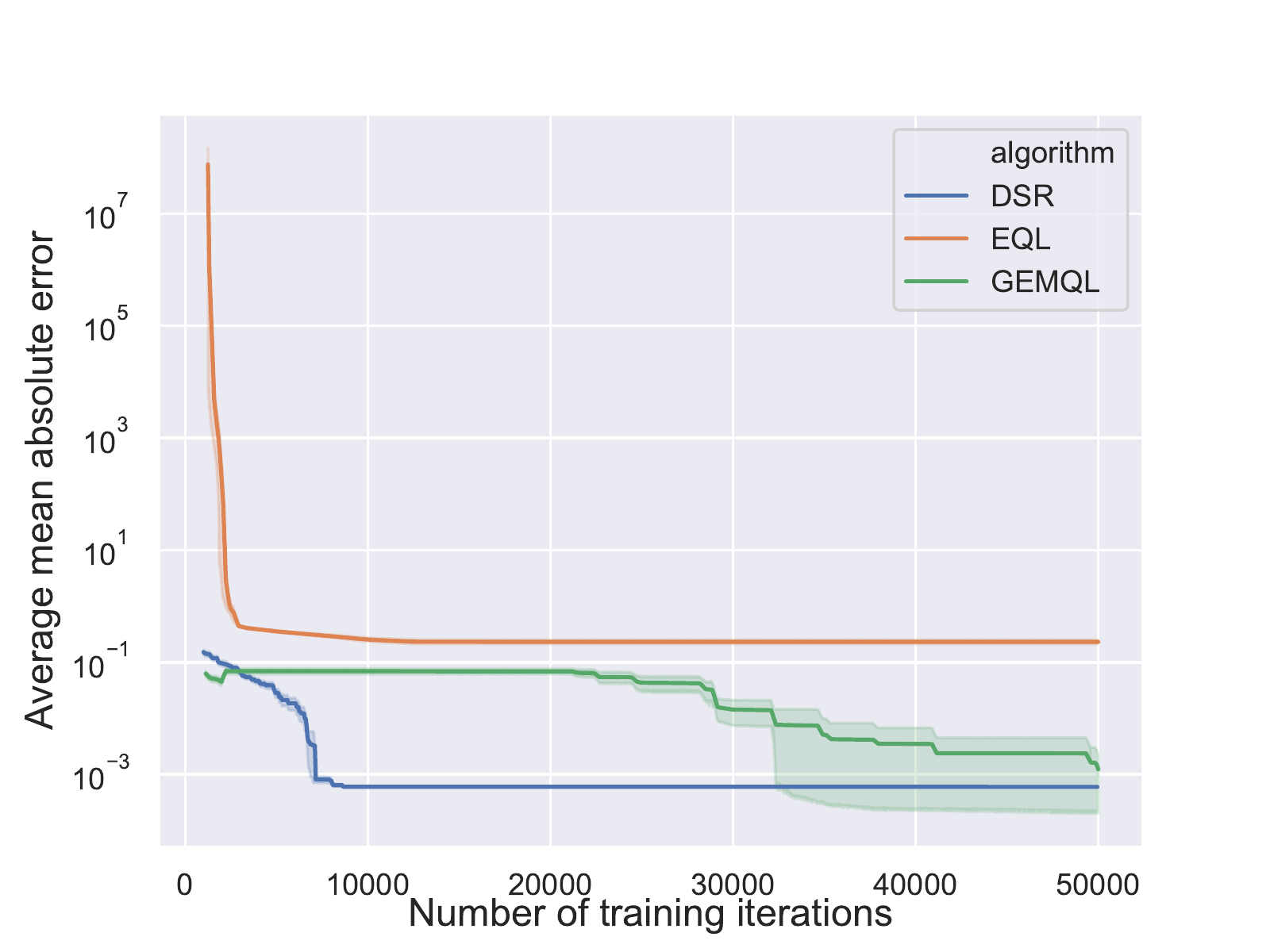}
\subcaption{$b6$}
\end{minipage}
\begin{minipage}[t]{0.24\textwidth}
\includegraphics[width=\textwidth]{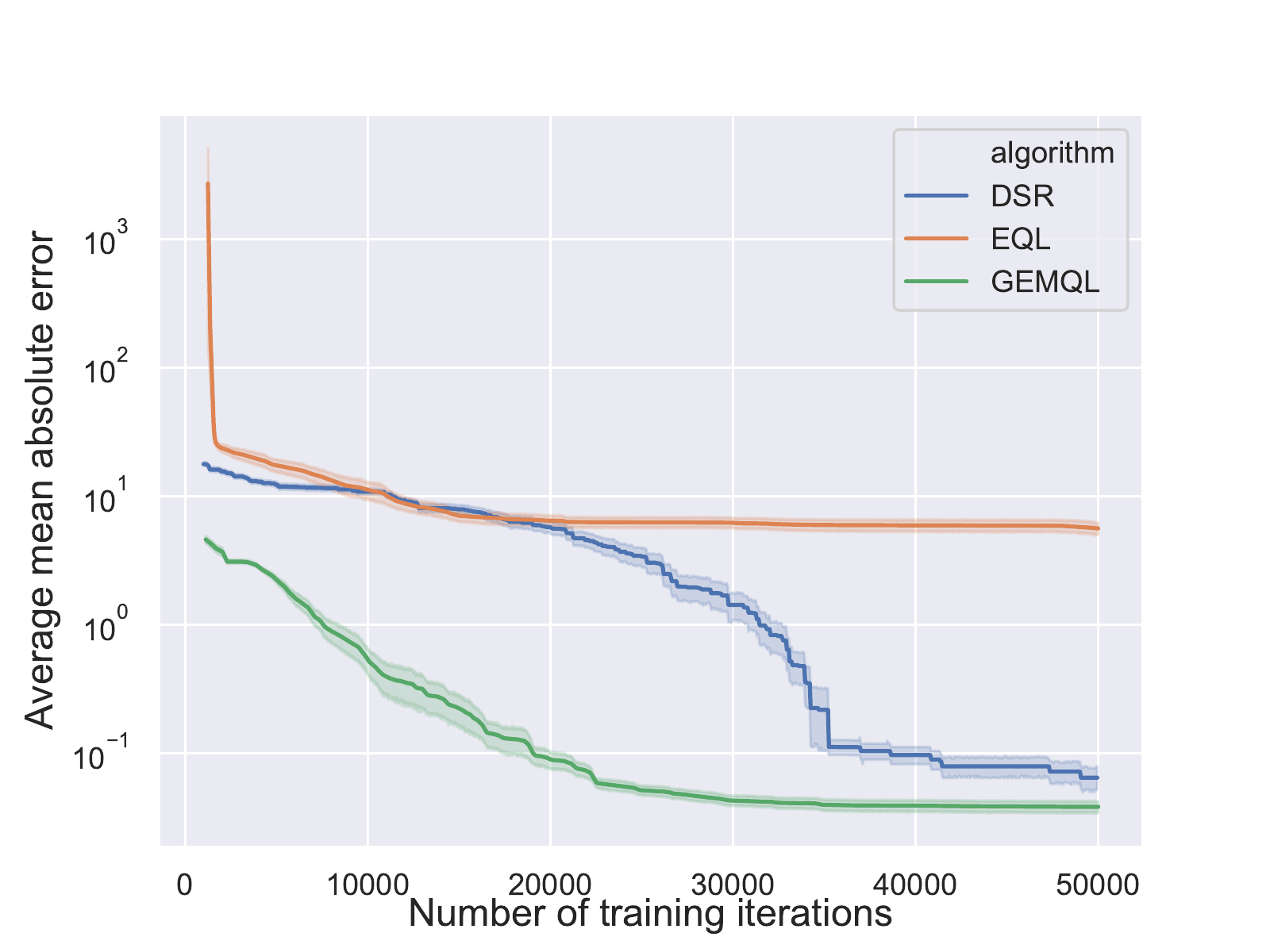}
\subcaption{$b7$}
\end{minipage}
\begin{minipage}[t]{0.24\textwidth}
\includegraphics[width=\textwidth]{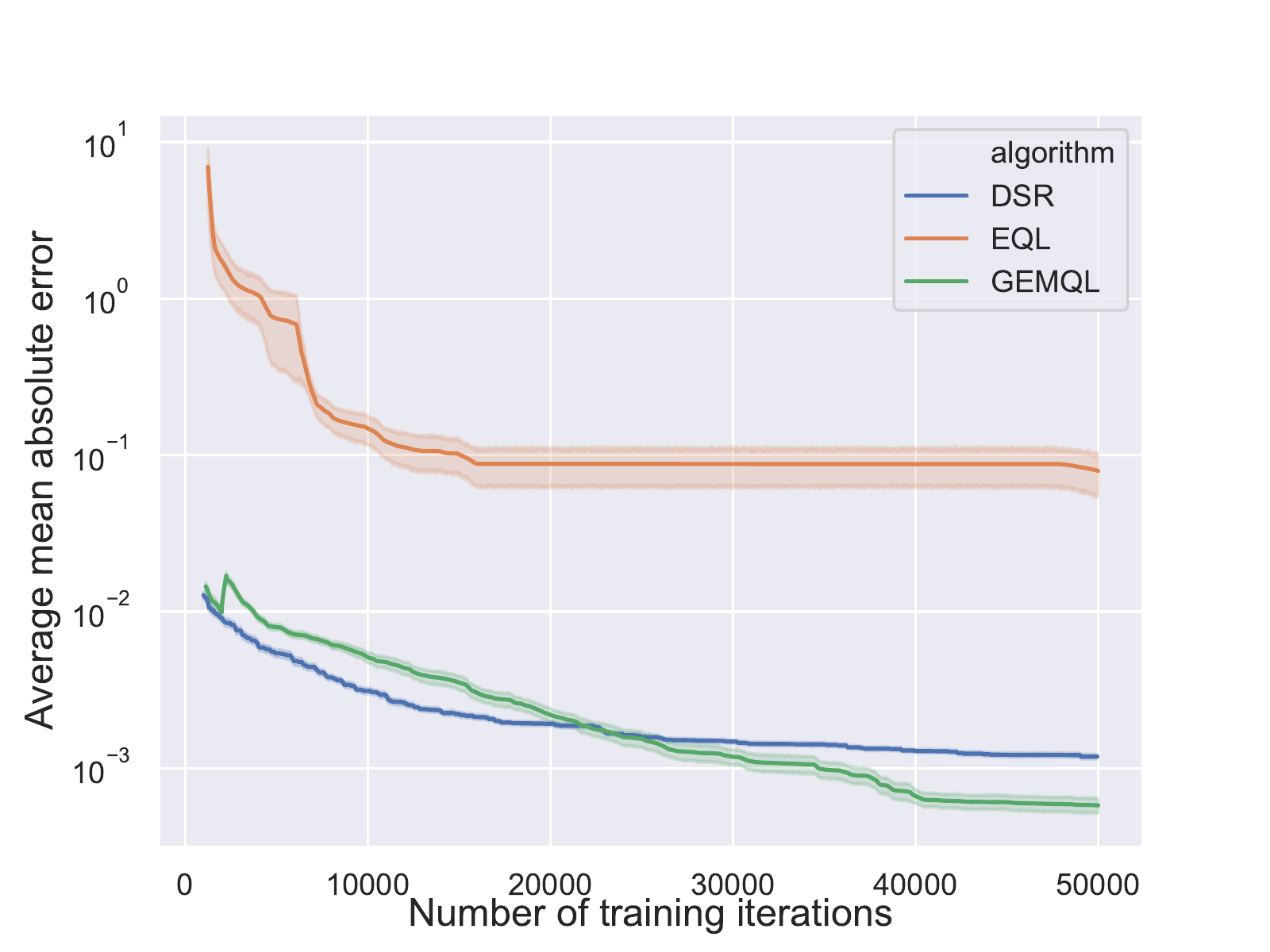}
\subcaption{$b8$}
\end{minipage}
\\
\begin{minipage}[t]{0.24\textwidth}
\includegraphics[width=\textwidth]{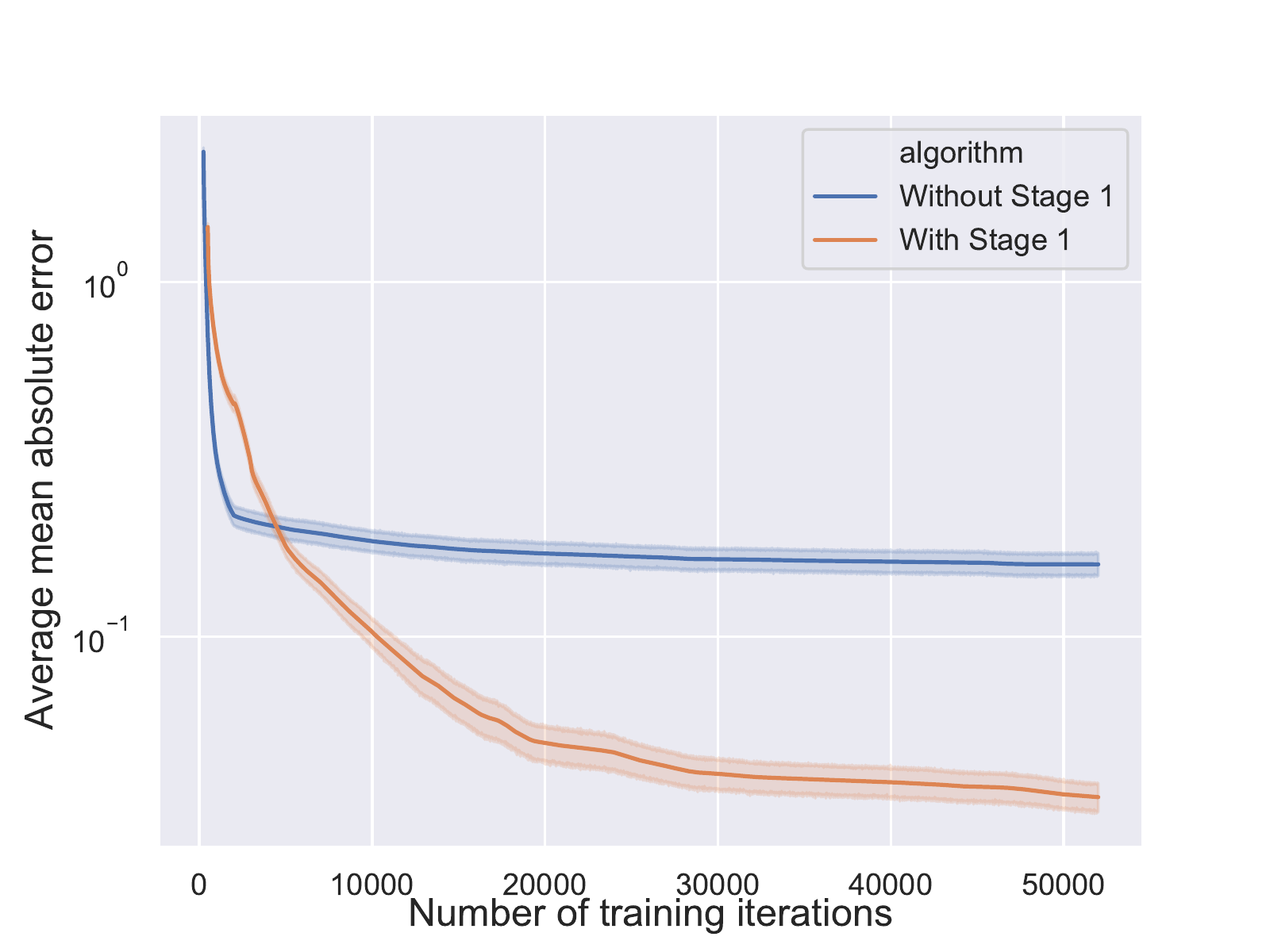}
\subcaption{Effect of stage 1 on $b4$}
\end{minipage}
\begin{minipage}[t]{0.24\textwidth}
\includegraphics[width=\textwidth]{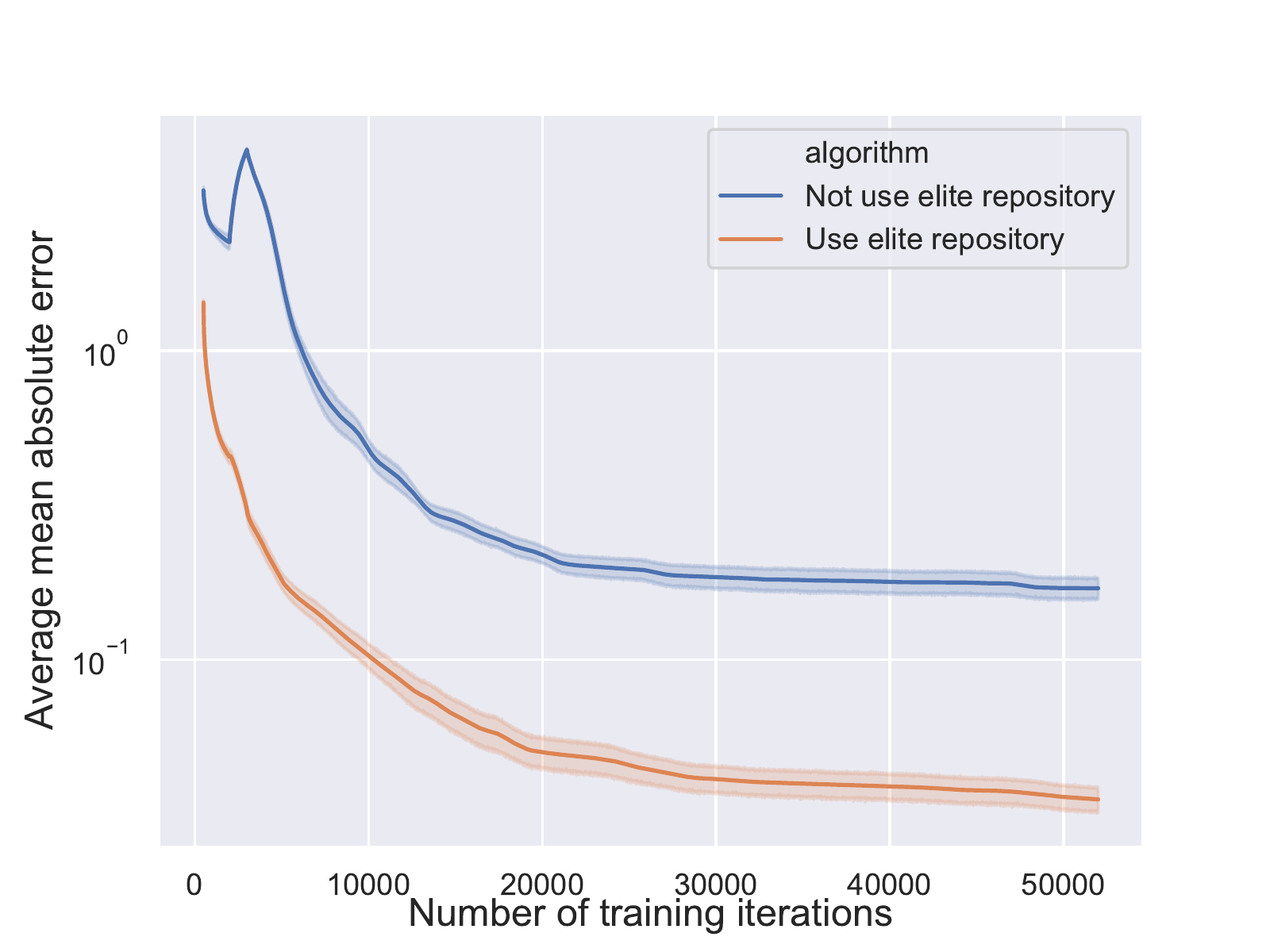}
\subcaption{Effect of using elite repository to sample network instances on $b4$}
\end{minipage}
\begin{minipage}[t]{0.24\textwidth}
\includegraphics[width=\textwidth]{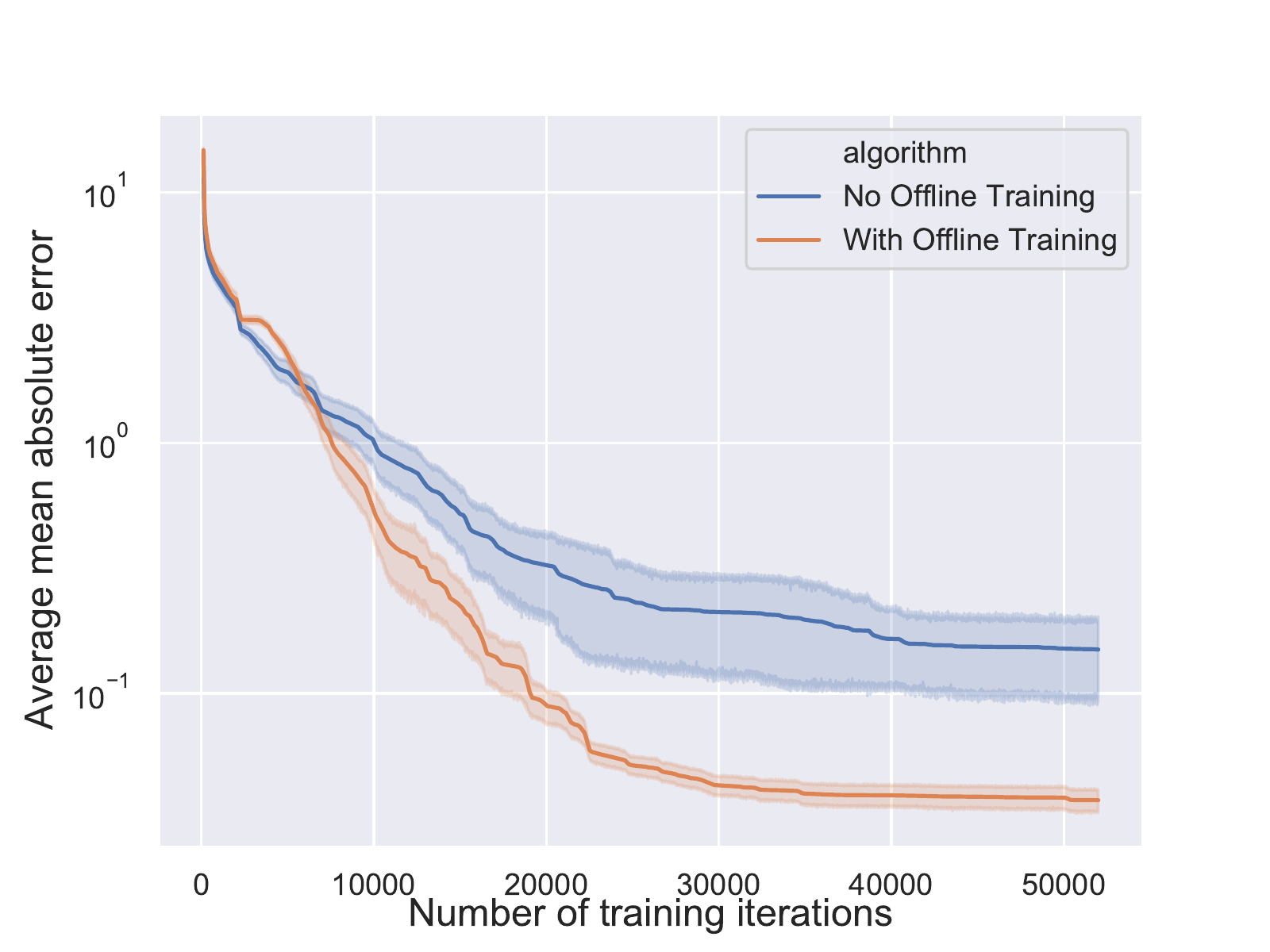}
\subcaption{Effect of offline training on $b7$}
\end{minipage}
\begin{minipage}[t]{0.24\textwidth}
\includegraphics[width=\textwidth]{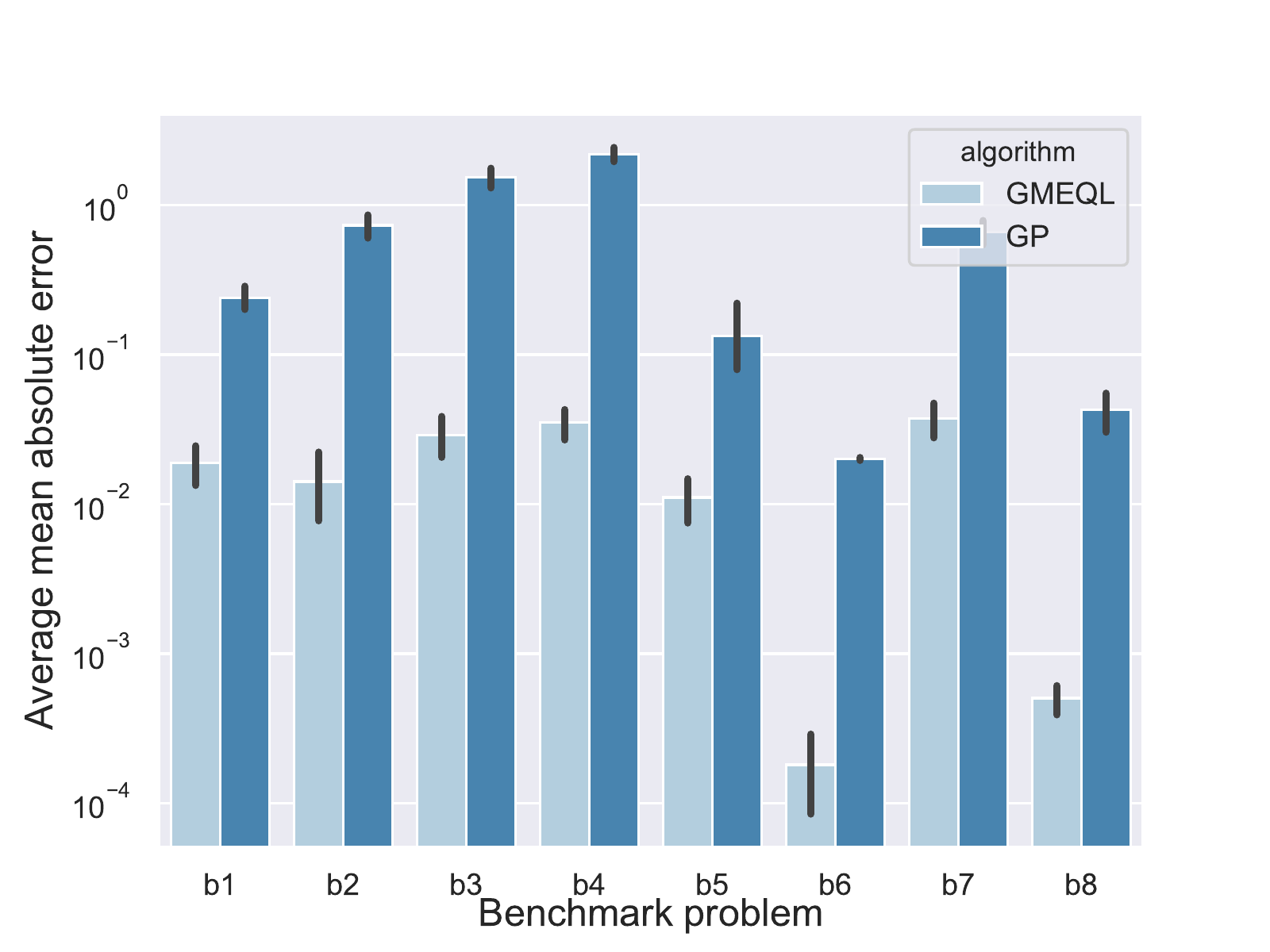}
\subcaption{Performance comparison with GP}
\end{minipage}
\caption{Performance evaluation of all competing algorithms on 8 benchmark problems.}
\label{fig-results}
\end{figure*}

GMEQL\footnote{https://github.com/aaron-vuw/gmeql} and EQL are implemented using TensorFlow\footnote{https://www.tensorflow.org/}. We also use the implementation of DSR provided by the algorithm inventors\footnote{https://github.com/brendenpetersen/deep-symbolic-regression}. The Python library GPlearn\footnote{https://gplearn.readthedocs.io/en/stable/intro.html} is used for all experiments on GP. We follow closely the hyper-parameter settings recommended in \cite{kim2020,petersen2019,koza1992} for EQL, DSR and GP. We set the population size to 5000 and the maximum number of generations to 100 for GP such that the total number of expressions to be evaluated during a GP run is at the same level as other competing algorithms. For GMEQL, we use Adam with default learning rate of 0.001 to train all network parameters. For other hyper-parameters introduced in Algorithm \ref{alg-train}, $n=3$, $m=2000$, $p=40$, $q=48000$ and $r=40$. Meanwhile, the size of the elite repository $\Delta$ is capped at 400. The temperature $\lambda=2/3$, following \cite{maddison2016}. We did not spend huge efforts to fine-tune any of these hyper-parameters. Small changes to $p$, $q$ and $r$ do not have noticeable impact on performance. By incrementing $n$ and $m$, more reliable learning behavior can be achieved at the expense of increased computation cost.

The main experiment results have been presented in Figure \ref{fig-results} (see Appendix C for more results). Each performance curve in this figure shows the average performance and STD of the respective algorithm over 20 independent runs. As evidenced in the first two rows of Figure \ref{fig-results}, GMEQL has clearly outperformed EQL and DSR on most of the benchmark problems. On problems $b6$ and $b7$, GMEQL and DSR achieved similar performance while DSR obtained slightly better accuracy on $b6$. However the performance difference on $b6$ is not significant. Our experiments suggest that learning symbolic expressions embedded in GMEQL directly can often produce more accurate results, in comparison to an indirect approach based on RNNs and reinforcement learning. Meanwhile, GMEQL achieved significantly higher regression accuracy than EQL.

We also perform ablation studies on the key components of Algorithm \ref{alg-train}. As shown in Figure \ref{fig-results}(i), without stage 1, the sampling of network instances and the training of structure parameters is strongly biased by the trained regression parameters. Despite of fast learning progress initially, subsequent training of GMEQL can easily converge to poor local optima. As indicated in Figure \ref{fig-results}(j), without using the elite repository, the chances of sampling accurate network structures can be substantially reduced, leading to clearly inferior performance. The temporary upward spike noticeable at the initial training phase in Figure\ref{fig-results}(j) is due to the shift from stage 1 to stage 2 in Algorithm \ref{alg-train} when all structure and regression parameters are reinitialized. Figure \ref{fig-results}(k) confirms that the offline training technique developed in Subsection \ref{sub-sec-offt} can effectively improve the accuracy of trained GMEQL and the sample efficiency of the training process.

GMEQL is compared with the standard GP method in Figure \ref{fig-results}(l). On all benchmark problems, GMEQL achieved clearly better regression accuracy. We have also checked the final expressions produced by GMEQL from every algorithm run. The success rates for GMEQL to find the ground-truth expressions (or mathematically equivalent expressions) are $61\%$, $70\%$, $45\%$, $40\%$, $53\%$, $100\%$, $47\%$ and $100\%$ with respect to $b1$ to $b8$ listed in Table \ref{tab-ben}. The observed success rates are consistently higher than other competing algorithms.

\section{Conclusion}
\label{sec-con}

In this paper we proposed a new NN architecture called GMEQL for symbolic regression. GMEQL applies continuous relaxation to the network structure via the Gumbel-Max trick and introduces two types of trainable parameters: structure parameters and regression parameters. To achieve high training performance and reliability, a new two-stage training process has been developed in this paper. We further proposed to use an elite repository to increase the chances of discovering accurate network samples. A new training technique in the offline setting has also been developed to expedite the training process and reduce the amount of network samples required for symbolic regression.

While GMEQL was mainly studied for symbolic regression tasks in this paper, we have performed some preliminary study of using GMEQL to find symbolic solutions of differential equations with initial success. Furthermore, GMEQL can potentially help to learn symbolic rules for robotics control, distributed resource allocation and other challenging decision-making problems. All these possibilities deserve substantial investigations in the future.

\bibliographystyle{named}
\bibliography{ijcai21}

\section*{Appendix A}

In this appendix, we analyze the probability for any dimension of vector $V_{i,j}^k$ in (\ref{equ-vij}) to be 1 (or close to 1), when temperature $\lambda$ is small. For any $1\leq l\leq M_{k-1}$, if
\begin{equation}
V_{i,j}^k(l)=\frac{\exp((\log\alpha_{i,j,l}^k+G_{i,j,l}^k)/\lambda)}{\sum_{p=1}^{M_{k-1}} \exp((\log\alpha_{i,j,p}^k+G_{i,j,p}^k)/\lambda)}
\label{equ-vij-gumbel}
\end{equation}
then, according to Proposition 1 in \cite{maddison2016}, the two statements below can be established:
$$
S1: \mathbb{P}(V_{i,j}^k(l)>V_{i,j}^k(p), \forall p\neq l)=\frac{\alpha_{i,j,l}^k}{\sum_{p=1}^{M_{k-1}}\alpha_{i,j,p}^k}
$$
$$
S2: \mathbb{P}(\lim_{\lambda\rightarrow\infty}V_{i,j}^k(l)=1)=\frac{\alpha_{i,j,l}^k}{\sum_{p=1}^{M_{k-1}}\alpha_{i,j,p}^k}
$$
Statement $S1$ is known as the \emph{rounding} statement. Statement $S2$ is the so-called \emph{zero temperature} statement. They together imply that, when $\lambda$ is small, the probability for the $l$-th dimension of $V_{i,j}^k$ to be 1 (or close to 1) is proportional to $\alpha_{i,j,l}$. Comparing (\ref{equ-vij}) and (\ref{equ-vij-gumbel}), by letting
\begin{equation}
\alpha_{i,j,l}^k=\exp(z_{i,j,l}^k),
\label{equ-alpha-z}
\end{equation}
it is straightforward to show that $V_{i,j}^k$ defined in (\ref{equ-vij}) is equivalent to $V_{i,j}^k$ defined in (\ref{equ-vij-gumbel}). Hence, following statements $S1$ and $S2$ above, it can be concluded that
$$
\mathbb{P}(V_{i,j}^k(l)\approx 1)=\frac{\exp(z_{i,j,l}^k)}{\sum_{l=1}^{M_{k-1}} \exp(z_{i,j,l}^k) }.
$$

\section*{Appendix B}

In this appendix, we investigate in more details how to calculate the gradient $\nabla_{z_{i,j,l}^k} \log \mathbb{P}_{\{Z_{i,j}^k\}}(T)$ with respect to any sampled network instance $T$ and one of its connection instances $c_{i,j,l}^k$. We will also develop a new approach for training structure parameters in the offline setting.

Because the connection instances for each connection of GMEQL are sampled independently, it is eligible to consider every connection $C_{i,j}^k$ in isolation. Therefore,
$$
\nabla_{z_{i,j,l}^k}\log \mathbb{P}_{\{Z_{i,j}^k\}}(T)=\nabla_{z_{i,j,l}^k}\log \mathbb{P}_{\{Z_{i,j}^k\}}(V_{i,j}^k=X_{i,j}),
$$
with $X_{i,j}=[x_{i,j,1},\ldots,x_{x,j,M_{k-1}}]^T$ representing the sampled degrees of involvement of all connection instances $c_{i,j,l}^k$ belonging to connection $C_{i,j}^k$ in $T$.
Following Definition 1 in \cite{maddison2016}, if $V_{i,j}^k$ is determined randomly according to (\ref{equ-vij-gumbel}), then the \emph{probability density} for $V_{i,j}^k=X_{i,j}$ is:
\begin{equation}
\begin{split}
&\mathbb{P}(V_{i,j}^k=X_{i,j})\\
&=(M_{k-1}-1)!\lambda^{M_{k-1}-1} \prod_{a=1}^{M_{k-1}}\left( \frac{\alpha_{i,j,a}^k x_{i,j,a}^{-\lambda-1}}{\sum_{b=1}^{M_{k-1}}\alpha_{i,j,b} x_{i,j,b}^{-\lambda}} \right)
\end{split}
\label{equ-vij-density}
\end{equation}
Using (\ref{equ-alpha-z}),
\begin{equation*}
\begin{split}
& \nabla_{z_{i,j,l}^k} \log \mathbb{P}_{\{Z_{i,j}^k\}}(V_{i,j}^k=X_{i,j})= \\
& \sum_{a=1}^{M_{k-1}} \nabla_{z_{i,j,l}^k} \log\left( \frac{\exp(z_{i,j,a}^k) x_{i,j,a}^{-\lambda-1}}{\sum_{b=1}^{M_{k-1}} \exp(z_{i,j,b}^k) x_{i,j,b}^{-\lambda}} \right)
\end{split}
\end{equation*}
Consequently, we can derive the gradient calculation formula presented in (\ref{equ-grad-form}).

In line with (\ref{equ-max-l}) and (\ref{equ-g}), when $\{Z_{i,j}^k\}$ in (\ref{equ-max-l}) approach their optimal values, the gradient computed in (\ref{equ-g}) and subsequently in (\ref{equ-grad-form}) is expected to be very close to 0. Taking this reasoning one step further and considering any specific network instance $T$ sampled according to $\mathbb{Q}$, if $\nabla_{z_{i,j,l}^k} \log \mathbb{P}_{\{Z_{i,j}^k\}}(T)$ in (\ref{equ-g}) is close to 0, it is likely for $\{Z_{i,j}^k\}$ to be close to its optimal values (at least local optimal values) in (\ref{equ-max-l}). Driven by this idea, it can be easily verified from (\ref{equ-grad-form}) that if
\begin{equation}
\exp(z_{i,j,l}^k)=\left(V_{i,j}^k(l)\right)^{\lambda}, \forall i,j,k,l,
\label{equ-z-new}
\end{equation}
then
\begin{equation*}
\begin{split}
&\nabla_{z_{i,j,l}^k} \log \mathbb{P}_{\{Z_{i,j}^k\}}(T)\\
&=\frac{\sum_{a=1,a\neq l}^{M_{k-1}}1}{M_{k-1}}-\frac{M_{k-1}-1}{M_{k-1}} \\
&=0.
\end{split}
\end{equation*}
In view of the above, structure parameters $\{Z_{i,j}^k\}$ can be trained in the offline setting to minimize $J'$ below:
$$
J'=\frac{1}{\|\mathcal{B}\|}\sum_{T\in\mathcal{B}} \sum_{i,j,k,a} \left( \exp(z_{i,j,a}^k) - (V_{i,j}^k(a))^{\lambda} \right)^2
$$
with respect to a batch $\mathcal{B}$ of network instances sampled according to $\mathbb{Q}$ from the elite repository $\Delta$. In other words, every structure parameter $z_{i,j,l}^k$ can be updated along the gradient direction -$\nabla_{z_{i,j,l}^k}J'$. We have experimentally evaluated this approach for offline training of $\{Z_{i,j}^k\}$. Our experiments (see Appendix C) produced similar performance results, in comparison to training $\{Z_{i,j}^k\}$ based on gradients computed in (\ref{equ-grad-form}).

We therefore conclude that multiple different learning rules can be derived from the lower bound in (\ref{equ-j-bound}) to train $\{Z_{i,j}^k\}$ offline. In fact, the (local) optimal values of $\{z_{i,j,l}^k\}_{l=1}^{M_{k-1}}$ with respect to (\ref{equ-max-l}) is unlikely to be unique. Besides (\ref{equ-z-new}), it is possible to find other values of $\{z_{i,j,l}^k\}_{l=1}^{M_{k-1}}$ such that $\nabla_{z_{i,j,l}^k} \log \mathbb{P}_{\{Z_{i,j}^k\}}(T)=0$. Among all the alternatives, (\ref{equ-z-new}) can be easily derived from (\ref{equ-grad-form}) and is efficient to compute. We will not explore other (local) optimal values of $\{z_{i,j,l}^k\}_{l=1}^{M_{k-1}}$ in this paper.

\section*{Appendix C}

In this appendix, we first introduce the hardware/software settings of all experiments. We then discuss the average running time of all competing algorithms. Afterwards, the effectiveness of using (\ref{equ-z-new}) to train structure parameters in the offline setting will be experimentally evaluated. Finally, we will investigate the reliably of GMEQL at handling noisy datasets.

All experiments were carried out using Python code. The Python implementation of GMEQL has been made publicly accessible\footnote{https://github.com/aaron-vuw/gmeql}. To run the Python code, we have the Python 3.6.10 software installed on every computer utilized for the experiments. We also installed TensorFlow (1.14.0), Numpy (1.19.3) and Matplotlib (3.0.2) libraries as requested by the Python code.

No special hardware is needed to run our Python code. We used a group of commodity-grade desktop computers to perform all the experiments reported in the paper. Each computer is installed with the Arch Linux operating system. They have the same/similar hardware configurations, featuring particularly Intel i7 series of processors (both 7-th and 8-th generations) and 8GB of physical memory. No GPUs were involved in any experiments. Each algorithm was configured to run exclusively on one processor core.

In terms of the algorithm running time, GMEQL requires more running time than other algorithms. The average running time of GMEQL on a benchmark problem is about 3 to 4 hours. In comparison, EQL and GP can finish execution within 30 minutes. DSR requires about 2 hours to complete one run. It is to be noted that the observed algorithm running time can be substantially shortened by using multiple CPU cores or GPU acceleration techniques. We do not consider the difference in running time as a major drawback of GMEQL. However, highly efficient implementation of GMEQL and the corresponding training algorithms on massively parallel computing infrastructures is beyond the scope of this paper.
\begin{figure}[!ht]
\centering
\includegraphics[width=0.7\columnwidth]{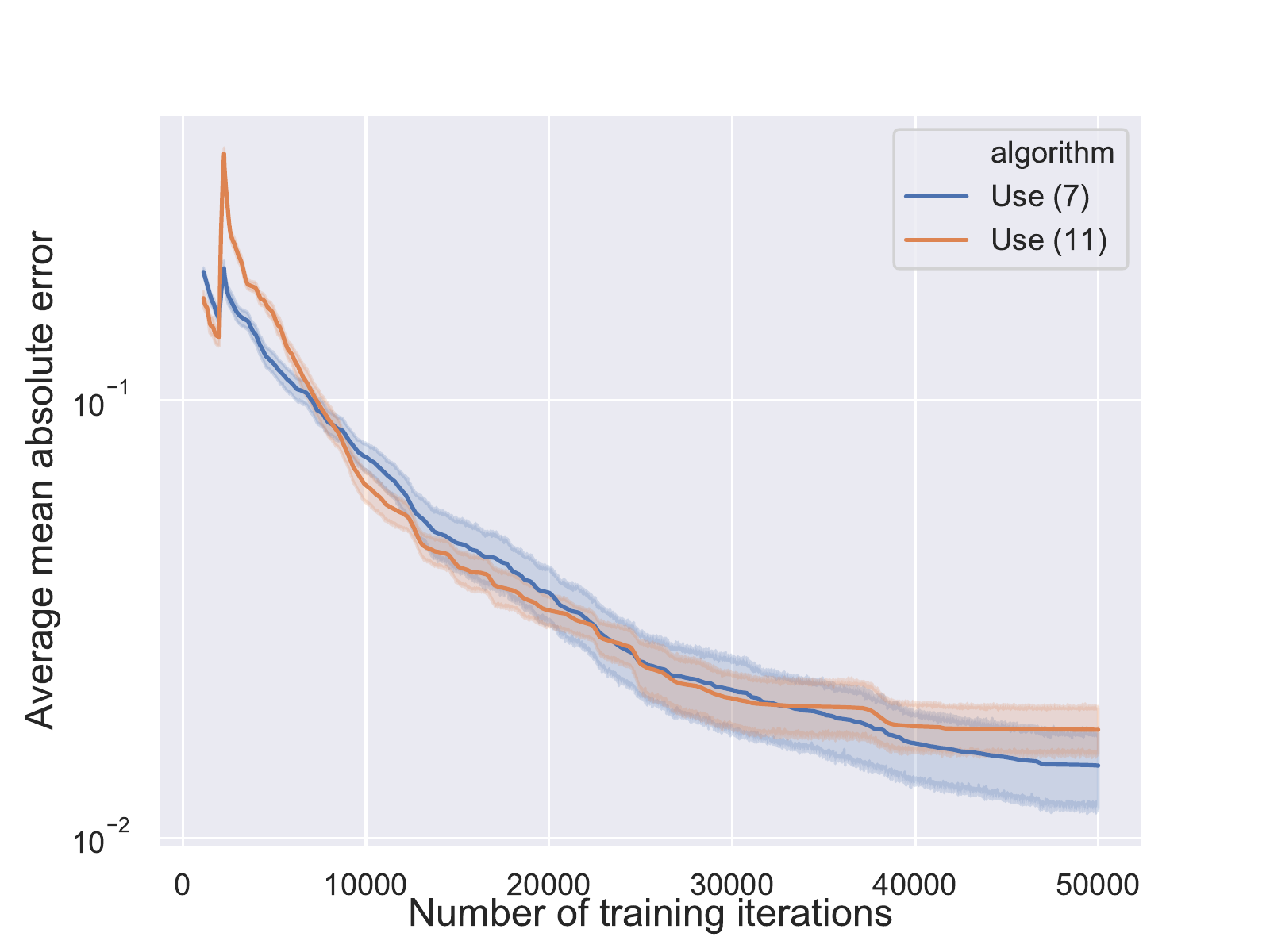}
\caption{Performance difference of using (\ref{equ-grad-form}) and (\ref{equ-z-new}) to train structure parameters offline on $b2$.}
\label{fig-offline-comp}
\end{figure}

Figure \ref{fig-offline-comp} compares the performance differences upon using (\ref{equ-grad-form}) and (\ref{equ-z-new}) respectively to train structure parameters in the offline setting. According to this figure, the performance differences are not statistically significant. This suggests that the offline learning rules for training structure parameters are not unique. Regardless of which rule is used, they can effectively improve the sample efficiency of the training process, as evidenced in Figure \ref{fig-results}(k). While Figure \ref{fig-offline-comp} focuses only on the benchmark problem $b2$, similar observations have also been witnessed on other benchmark problems.

\begin{figure}[!ht]
\centering
\includegraphics[width=0.7\columnwidth]{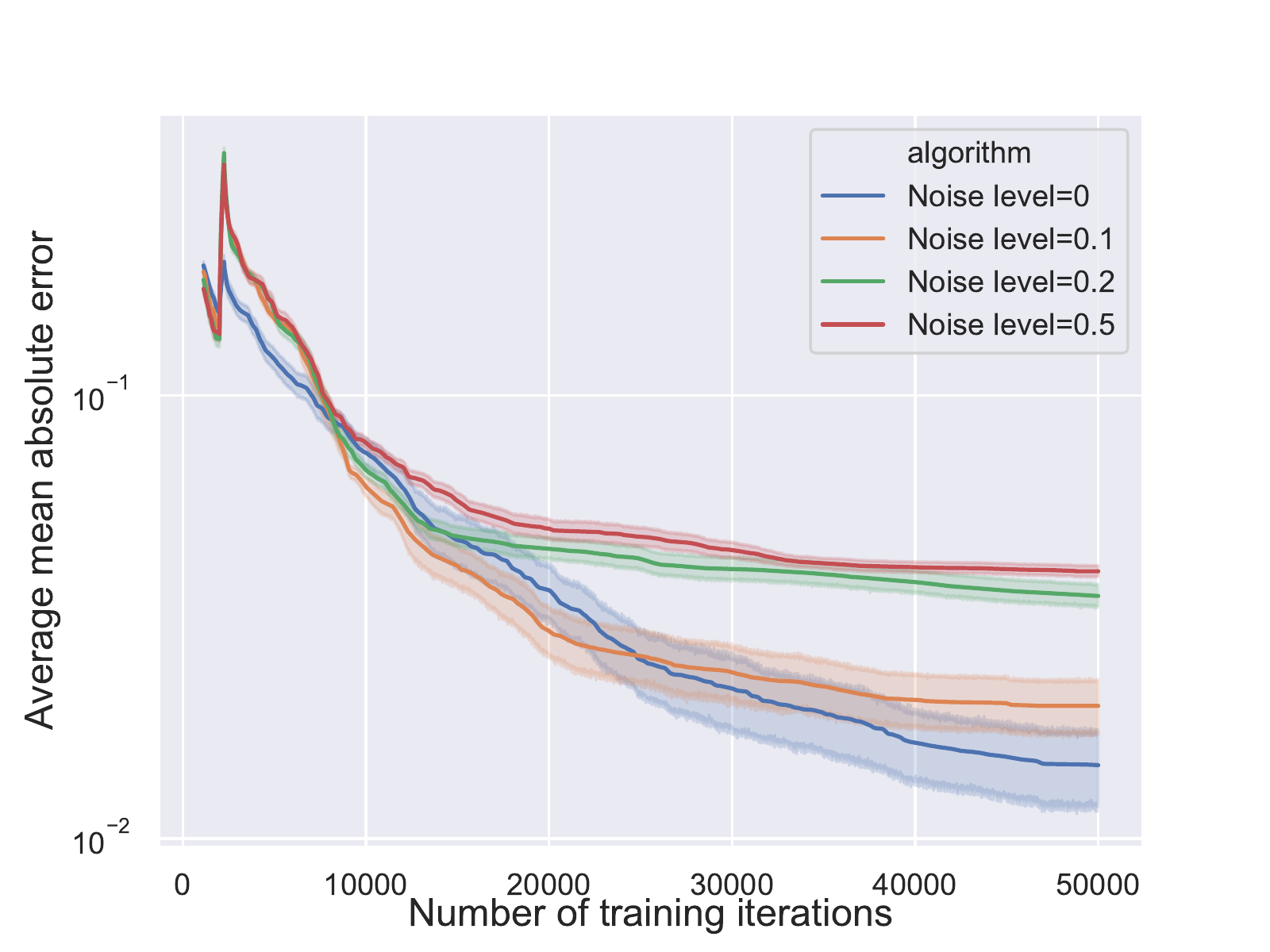}
\caption{Impact of different noise levels on the performance of GMEQL on $b2$.}
\label{fig-noise-comp}
\end{figure}

Experiment results in Figure \ref{fig-results} and Figure \ref{fig-offline-comp} were obtained by using noise-free datasets. In practice, the dataset for symbolic regression may be subject to certain levels of noises. To understand the performance impact of noise on GMEQL, we have tested the noisy version of each benchmark problem. Specifically, we add additional Gaussian noise with zero mean to the target output of each training instance in the benchmark dataset. The corresponding noise level is determined by the STD of the Gaussian noise. Different from training instances, all testing instances are noise-free in order to determine the true accuracy of GMEQL. Hence, for each benchmark problem, there are 300 noisy training instances and 300 noise-free testing instances. All accuracy results reported in Figure \ref{fig-noise-comp} were obtained by applying the mathematical expression with the best training accuracy discovered by GMEQL to all testing instances.

As shown in Figure \ref{fig-noise-comp}, when the noise level is not high (i.e. 0.1), GMEQL can achieve similar accuracy as the noise-free problem. The performance difference is negligible. On the other hand, upon increasing the noise level to 0.5, the accuracy difference becomes noticeable. Nevertheless, GMEQL still managed to achieve reasonably high accuracy. Across all the different noise levels examined in Figure \ref{fig-noise-comp}, GMEQL can successfully find the ground-truth expression during some algorithm runs. The observed success rates are 70\%, 65\%, 35\% and 32\% respectively, for noise levels of 0, 0.1, 0.2 and 0.5. In addition to the benchmark problem $b2$ studied in Figure \ref{fig-noise-comp}, similar results have also been obtained on other benchmark problems. In general, GMEQL can effectively cope with low levels of noise. When the noise level becomes high, additional noise reduction techniques may be required to maintain the accuracy of GMEQL. This will be studied in the future work.

\end{document}